\pdfoutput=1
\documentclass{article}
\usepackage{spconf,amsmath,graphicx}
\usepackage{color}
\usepackage{multirow}
\usepackage{balance}
\usepackage{fancyhdr}
\newcommand{\newvspace}{\vspace{-0.20cm}}
\newcommand{\imagevspace}{\vspace{-0.35cm}}

\title{ADVANCED LSTM: A STUDY ABOUT BETTER TIME DEPENDENCY MODELING IN EMOTION RECOGNITION}
\makeatother
\name{{\em Fei Tao$^{1\ast}$ \thanks{$\ast$This work was done during the author's summer internship at Alibaba Group (U.S.) Inc.}, Gang Liu$^2$\vspace{-0.4cm}}}
\address{1. Multimodal Signal Processing (MSP) Lab, The University of Texas at Dallas, Richardson TX\\
	2. Institute of Data Science and Technology (iDST)-Speech, Alibaba Group (U.S.) Inc.\\
	{\footnotesize \tt \hspace{-0.1cm}fxt120230@utdallas.edu,g.liu@alibaba-inc.com} \vspace{-0.5cm}}
\begin{document}
%
%
\maketitle
\begin{abstract}
\emph{Long short-term memory} (LSTM) is normally used in \emph{recurrent neural network} (RNN) as basic recurrent unit. However, conventional LSTM assumes that the state at current time step depends on previous time step. This assumption constraints the time dependency modeling capability. In this study, we propose a new variation of LSTM, \emph{advanced LSTM} (A-LSTM), for better temporal context modeling. We employ A-LSTM in weighted pooling RNN for emotion recognition. The A-LSTM outperforms the conventional LSTM by 5.5\% relatively. The A-LSTM based weighted pooling RNN can also complement the state-of-the-art emotion classification framework. This shows the advantage of A-LSTM.
\end{abstract}

\imagevspace
\section{Introduction}
\label{sec:intro}
\newvspace
\emph{Recurrent neural network} is recently used as a dynamic model for sequential input. \emph{Long short-term memory} (LSTM) is usually adopted as basic units in RNN because it is able to solve the gradients vanishing and exploding problems in RNN training \cite{hochreiter_1997}. It uses memory cell and gates to control whether information will be memorized, output or forgotten. The LSTM takes two inputs, output from lower layer and output from previous time step in current layer. This configuration implies an assumption that the current state depends on the state of previous time step. This assumption of time dependency may constraint the modeling capability of RNN. In this paper, we propose a new variation of LSTM, \emph{advanced LSTM} (A-LSTM), to address this issue. In A-LSTM, current state depends on multiple states of different time steps. This releases the constrains in conventional LSTM and provides better time dependency modeling capability.

This paper presents our early study on A-LSTM. We explore the modeling capability of A-LSTM in the application of voice-based emotion recognition. Recognizing emotion based on audio in real world will improve the user experience of voice-based \emph{artificial intelligent} (AI) product, like Siri, Alexa. The input voice to system in real application may contains long silence (or pause) or non-speech voice filler, the conventional low level statistics feature like \emph{Interspeech 2010 paralinguistic challenge feature set} (IS10) or GeMAPs \cite{Schuller_2010,florian_2015}, may be failed. Weighted pooling based on attention mechanism is an appealing solution for these cases \cite{mirsamadi_2017}, which relies on RNN. We built an attention based weighted pooling framework with multi-task learning for emotion recognition in this study. When we apply A-LSTM in this framework, it gains 5.5\% relative improvement compared with conventional LSTM.

The remaining of the paper is organized as following structure. Section \ref{sec:related} reviews previous work. Section \ref{sec:corpus} introduces the IEMOCAP corpus which is used in this paper. Besides, the acoustic feature extraction is also described. Section \ref{sec:proposed} describes the details of the proposed approach. Section \ref{sec:experiments} describes the experiments and analysis of results. Section \ref{sec:conclusion} concludes the work and leads to the future direction of the work.

\imagevspace
\section{Related Work}
\label{sec:related}
\newvspace
\cite{liu_2014} shows that temporal information is beneficial for emotion identification. \cite{snyder_2015,peddinti_2015,waibel_1989} shows that the performance of the neural network will be improved when higher layers can see more time steps from lower layer. These works rely on DNN rather than RNN. They do not discuss the timing sequence modeling. \cite{he_2016,srivastava_2015} proposed solutions to having alternative connections between layers in DNN. These solutions are different from the conventional connections within network. \cite{zhang_2016,kim_2017} modify the LSTM architecture relying on residual or highway connection. However, the modifications in these papers are focusing on connecting the memory cells between lower and higher layers. They do not modify the connection within the same layer. \cite{zhang_2015} modifies the output hidden value to higher layer by a weighted summation. \cite{soltani_2016} follows similar idea. It uses weighted pooling of the hidden values of multiple historic time steps at each time steps which improves the information richness to higher layer. This is equivalent to allow higher layer see more time steps. But they do not modify the memory cell which means the time dependency is not changed. \cite{wang_2016} shows that the combination of near time steps may not improve the system a lot. The combination should contain a long term range. Besides, they do not combine the multiple states at each step, which is different from the high order RNN.
For emotion recognition, \cite{mirsamadi_2017} recently proposed attention based weighted pooling RNN to extract acoustic representation. The work shows the weighted pooling RNN can outperform conventional pooling approach, like mean, maximum, or minimum. It also shows the RNN framework can capture the section of interest. Multi-task learning recently shows its advantage in emotion recognition task \cite{xia_2017,parthasarathy_2017_1}. But in these papers, the regression of valence and arousal values are normally set as auxiliary tasks, which is hard to obtain.

\imagevspace
\section{Corpus Description and Feature Extraction}
\label{sec:corpus}
\newvspace
We apply A-LSTM in the application of categorical emotion classification. We used IEMOCAP \cite{Busso_2008_5} corpus in this study which has 5 sections and 10 actors in total. In each section, there were two actors (one male and one female) involved in scripted or spontaneous scenarios to perform specific emotions. The utterances were segmented and with one categorical label, which is among angry, fear, excited, neutral, disgust, surprised, sad, happy, frustrated, other and XXX. XXX was the case that the annotators were not able to have agreement on the label. The corpus has 10039 utterances with average duration of 4.5 s per utterance (12.55 hr in total). The distribution of emotion classes is not balanced. In this study, we select 4 classes, neutral, happy, angry and sad. The total number of utterances used is 4490.

The corpus has video and audio channels. We only used audios in this study. The audio was collected by high quality microphones (Schoeps CMIT 5U) at the sample rate of 48 kHz. We downsampled them to 16 kHz and extract a 36D acoustic feature. The acoustic feature includes 13D MFCCs, \emph{zero crossing rate} (ZCR), energy, entropy of energy, spectral centroid, spectral spread, spectral entropy, spectral flux, spectral rolloff, 12D chroma vector, chroma deviation, harmonic ratio and pitch. The extraction was performed within a 25 ms window whose shifting step size was 10 ms (100 fps). The acoustic feature sequence was z-normalized within each utterance.

\imagevspace
\section{Proposed Approach}
\label{sec:proposed}
\newvspace
\subsection{Attention Based Weighted Pooling RNN}
\label{sec:weight_pool_rnn}
\newvspace
Attention based weighted pooling RNN is a data-driven framework to learn utterance representation from data, which can be suitable for practical application  \cite{mirsamadi_2017}. It relies on the attention mechanism \cite{bahdanau_2016} to learn the weight of each time step. The weighted summation is then computed as the representation of the whole utterance. Multi-task learning incorporates several aspects of knowledge into training, therefore it can learn better representation.

\begin{figure}[tb]
	\centering
	{
		\includegraphics[width=\columnwidth]{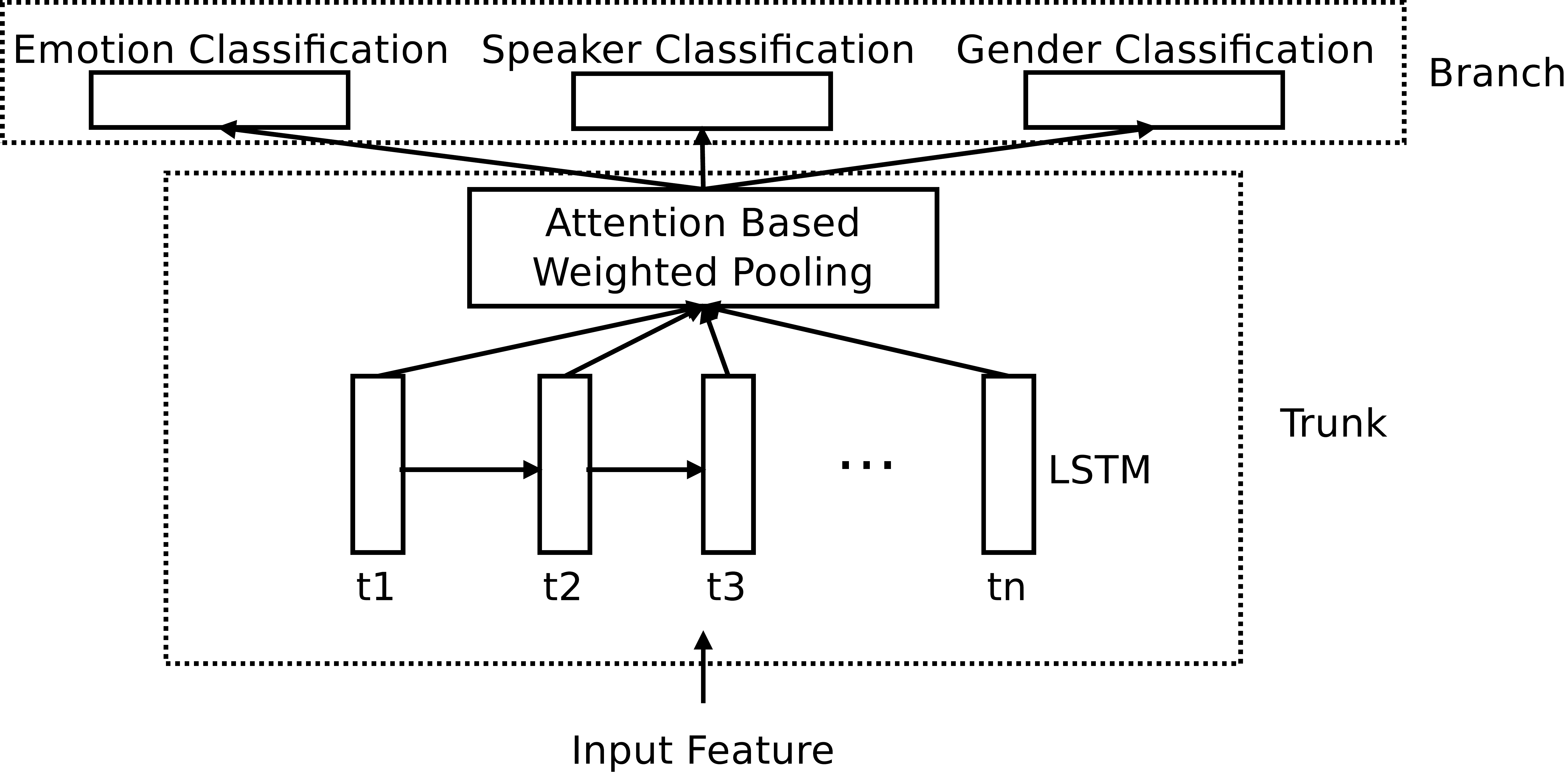}
	}\hspace{0.01mm}
	\imagevspace
	\caption{The attention based weighted pooling RNN. The LSTM layer is unrolled along the time axis (time t1 to tn). The trunk part has the layers that are shared by all the tasks. On top of the trunk part, there is branch part for tasks. The main task is emotion classification. The auxiliary tasks are speaker and gender classifications.}
	\newvspace
	\label{fig:rnn}
\end{figure}

The system diagram is shown in Figure \ref{fig:rnn}. The diagram has two parts, trunk and branch (two dashed boxes in the diagram). The branch is the part for different tasks, which includes emotion, speaker and gender classifications in this study. The trunk is the shared part of all tasks. The attention based weighted pooling is computed as Equation \ref{eq:temp_pool}, where $h_T$ is the hidden value output from the LSTM layer at time $T$, and $A_T$ is a scalar number representing the corresponding weight at time $T$. $A_T$ is computed in a softmax fashion following Equation \ref{eq:temp_weight}, where $W$ is a parameter need be learned. $\exp(W \cdot h_T)$ represents the potential energy at time $T$. This is similar to attention mechanism. If the frame at time $T$ has high potential energy, its weight will be high and therefore gain high ``attention"; if the potential energy is low, the weight and ``attention" will also be low. By this way, the model can learn to assign weights to different time steps from data.

If weights at all time steps are same, the weighted pooling is equal to arithmetic mean.

\begin{equation} 
\label{eq:temp_pool}
Weighted Pooling = \sum_{T=t1}^{tn} A_T \times h_T
\vspace{-0.2cm}
\end{equation}

\begin{equation} 
\label{eq:temp_weight}
A_T = \dfrac{\exp(W \cdot h_T)}{\sum_{T=t1}^{tn}\exp(W \cdot h_T)}
\vspace{-0.2cm}
\end{equation}

In this study, we define that trunk part has two hidden layers. The first layer is fully connected layer which has 256 RELU neurons. The second one is a \emph{bidirectional LSTM} (BLSTM) layer with 128 neurons. The hidden values go to weighted pooling layer after the LSTM layer. In the branch part, each task has one hidden fully connected layer with 256 RELU neurons and one softmax layer performing classification.

\newvspace
\subsection{Advanced LSTM}
\begin{figure}[tb]
	\centering
	{
		\includegraphics[width=\columnwidth]{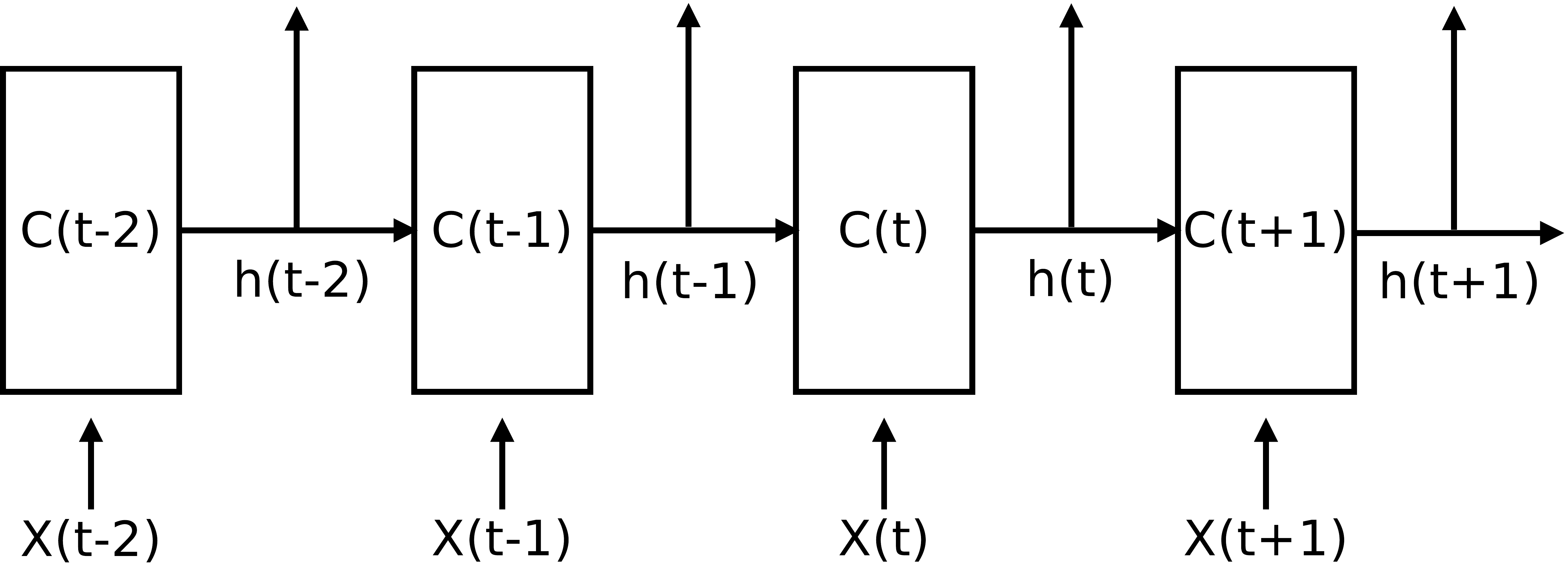}
	}\hspace{0.01mm}
	\imagevspace
	\caption{The unrolled conventional LSTM. Unrolling is along the time axis. The $C$ is the cell memory, $X$ is the values from lower layer, and $h$ is the hidden values to higher layer. States at time $t$ depends on the one at time $t-1$ in conventional LSTM.}
	\newvspace
	\label{fig:LSTM}
\end{figure}

\newvspace
Conventional LSTM tasks take the output from lower layer and previous time step as input and feed value to higher layer. The gating mechanism is used to control information flow by point-wise multiplication (denoted as $\odot$ operation in the following contents). There is a cell to memorize information within the unit. The diagram is shown in Figure \ref{fig:LSTM}. 

The cell is updated as Equation
\ref{eq:LSTM_cell}, where $f_t$ and $i_t$ are the forgetting and inputting gates at time $t$. $\widetilde{C}_t$ is new candidate cell values. It is computed as Equation \ref{eq:LSTM_cell_candidate}, where $tanh$ is the activation function, $W_C$ is a set of weights to be learned, $b_C$ is the bias, and $[h_{t-1},x_t]$ is the concatenation of the values from previous time step ($h$ value) and lower layer ($x$ value). $h$ value at time $t$ is computed by Equation \ref{eq:LSTM_h}, where $o_t$ is outputting gate. It can be seen that the states at time $t$ depends on the states at time $t-1$, because $C_t$ is computed from $h_{t-1}$ and $C_{t-1}$. The computation about controlling gates are omitted for simplification.

\begin{equation} 
\label{eq:LSTM_cell}
C_t = f_t \odot C_{t-1} + i_t \odot \widetilde{C}_t
\vspace{-0.2cm}
\end{equation}

\begin{equation} 
\label{eq:LSTM_cell_candidate}
\widetilde{C}_t = tanh(W_C \cdot [h_{t-1},x_t] + b_C)
\vspace{-0.2cm}
\end{equation}

\begin{equation} 
\label{eq:LSTM_h}
h_t = o_t \odot tanh(C_t)
\vspace{-0.2cm}
\end{equation}

\begin{figure}[tb]
	\centering
	{
		\includegraphics[width=\columnwidth]{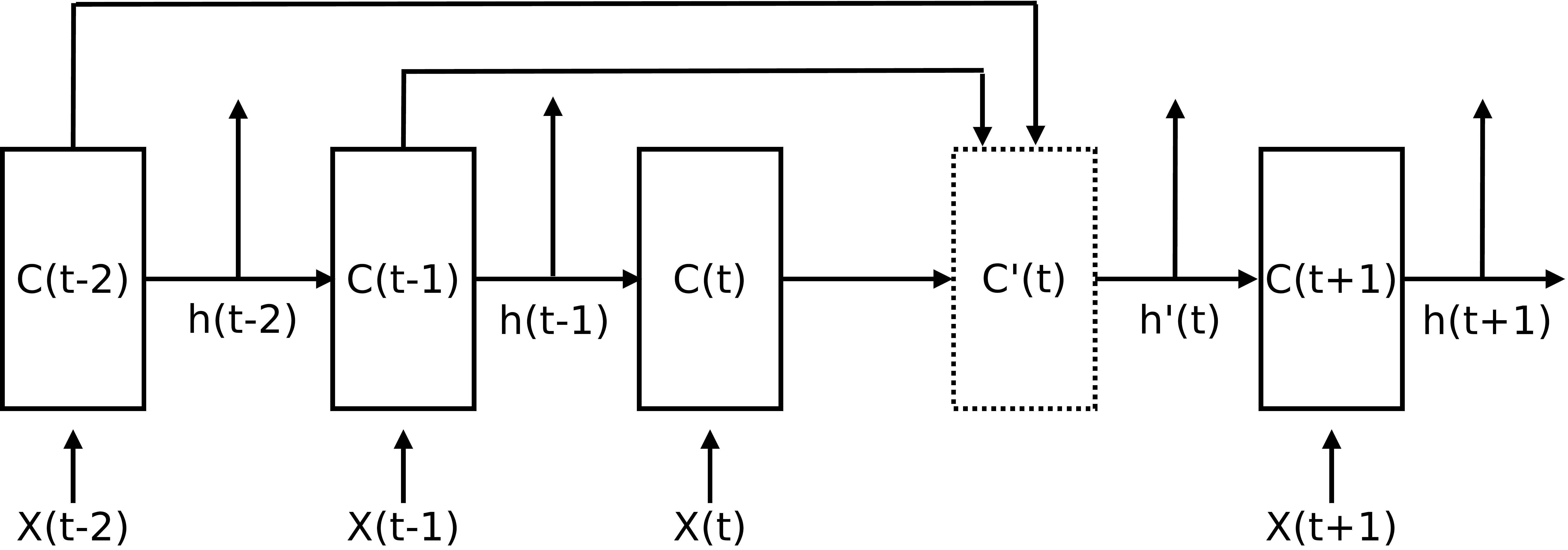}
	}\hspace{0.01mm}
	\imagevspace
	\caption{The unrolled A-LSTM. Unrolling is along the time axis. The $C$ is the cell memory, $X$ is the values from lower layer, and $h$ is the hidden values to higher layer. The dashed box is a weighted summation operation to combine the states at time $t-2$, $t-1$ and $t$. $C'$ and $h'$ is new cell memory and hidden value after combination. They are passed to compute the states at time $t+1$.}
	\newvspace
	\label{fig:A-LSTM}
\end{figure}

The A-LSTM is different from the conventional one. It releases the assumption that time $t$ state depends on time $t-1$ state. It use weighted summation of multiple states at different time steps to compute cell ($C$ value) and hidden value ($h$ value). The diagram is shown in Figure \ref{fig:A-LSTM}.

In A-LSTM, Equation \ref{eq:LSTM_cell} is modified to Equation \ref{eq:A-LSTM_cell}, and Equation \ref{eq:LSTM_cell_candidate} is modified to Equation \ref{eq:A-LSTM_cell_candidate}. $C'$ is computed following Equation \ref{eq:cell_update}, where $T$ is the set of selected time steps to be combined. In Figure \ref{fig:A-LSTM}, the selected time steps is $t-2$, $t-1$ and $t$ for time $t+1$. $T$ is therefore denoted as a set of \{3,2,1\}. In the remaining contents, we follow the same naming convention to show our configuration of A-LSTM. $W_{C_T}$ is a scalar number as corresponding weight at a specific time step. It is learned from Equation \ref{eq:A-LSTM_cell_weight}. Candidate value of $h$ at time $t$ is computed following Equation \ref{eq:A-LSTM_h_candidate}. It is same as Equation \ref{eq:LSTM_h} except the cell value now is updated to $C'$. After $h_t$ is obtained, the computation of $h'$ is computed following Equation \ref{eq:h_update} and \ref{eq:A-LSTM_h_weight}. The equations are similar to $C'$ computation. In Equation \ref{eq:A-LSTM_cell_weight} and \ref{eq:A-LSTM_h_weight}, $W$ is shared, which is the parameter to be learned from data. In this study, $C'$ and $h'$ are computed every $\max(T)$ steps rather than every step. For example, in the case of Figure \ref{fig:A-LSTM}, they are computed every 3 steps. 

A-LSTM is able to allow more flexible time dependency modeling capability. It makes the cell to recall far back historic records. Recalling every once in a while will be like the human learning mechanism, which makes learning better. Therefore the cell memory can  memorize information better compared with conventional LSTM.

\begin{equation} 
\label{eq:A-LSTM_cell}
C_t = f_t \odot C'_{t-1} + i_t \odot \widetilde{C}_t
\vspace{-0.2cm}
\end{equation}

\begin{equation} 
\label{eq:A-LSTM_cell_candidate}
\widetilde{C}_t = tanh(W_C \cdot [h'_{t-1},x_t] + b_C)
\vspace{-0.2cm}
\end{equation}

\begin{equation} 
\label{eq:cell_update}
C' = \sum_{T} W_{C_T} \times C_T
\vspace{-0.2cm}
\end{equation}

\begin{equation} 
\label{eq:A-LSTM_cell_weight}
W_{C_T} = \dfrac{\exp(W \cdot C_T)}{\sum_{T}\exp(W \cdot C_T)}
\vspace{-0.2cm}
\end{equation}

\begin{equation} 
\label{eq:A-LSTM_h_candidate}
h_t = o_t \odot tanh(C'_t)
\vspace{-0.2cm}
\end{equation}

\begin{equation} 
\label{eq:h_update}
h' = \sum_{T} W_{h_T} \times h_T
\vspace{-0.2cm}
\end{equation}

\begin{equation} 
\label{eq:A-LSTM_h_weight}
W_{h_T} = \dfrac{\exp(W \cdot h_T)}{\sum_{T}\exp(W \cdot h_T)}
\vspace{-0.2cm}
\end{equation}

\newvspace
\section{Experiments and Results}
\label{sec:experiments}
\newvspace
We evaluate our proposed A-LSTM on selected utterances from IEMOCAP corpus, which belonged to neutral, happy, angry and sad classes. We run two sets of experiments. In the first one, we compared different types of LSTMs. All the systems were based on weighted pooling RNN framework. In the second one we compared RNN framework with a \emph{deep neural network} (DNN) framework, which represents current state-of-the-art system on IEMOCAP. Multi-task learning was applied during all the systems. The weights for emotion, speaker and gender classification were 1, 0.3, 0.6 respectively. We randomly selected 1 male and 1 female as testing subjects. The data from other subjects were used as training data. 10\% of the training data was used as validation data to check whether we need early stopping. The early stopping criteria was that in continuous 3 epochs, the accuracy on the validation data was lower than the highest accuracy. 

\emph{Macro average F-score} (MAF) (also named as unweighted average F-score) \emph{macro average precision} (MAP) (also named as unweighted average precision) and accuracy were used as performance metrics. The metrics were computed with the open source tool, Scikit-learn \cite{scikit-learn}. Since the classes were imbalanced, we mainly rely on the MAF for performance evaluation.

\newvspace
\subsection{Weighted Pooling RNN Results}
\newvspace
We built up two baseline systems for comparison under the RNN framework. The first one used conventional LSTM. The second one used recurrent unit that was similar to the A-LSTM structure except that $W_{C_t}$ and $W_{h_t}$ were fixed. They were determined to same values which made the combination equivalent to arithmetic mean of the states at selected time steps. We therefore name it as ``mean LSTM". The proposed framework used A-LSTM as recurrent unit. The parameter details of the neural network has been described in Section \ref{sec:weight_pool_rnn}. Dropout was used in all the layers in the network except the attention based weighted pooling layer and the parameter of $W$ in Equation \ref{eq:A-LSTM_cell_weight} and \ref{eq:A-LSTM_h_weight}. The dropout rate was 0.5. The set of $T$ for A-LSTM used in this experiment was \{5, 3, 1\}. The time steps were selected every 2 time points. It was observed in pilot experiment that the training would be difficult when too many times steps in $T$, we therefore fixed to 3 selected time steps. Adam \cite{kingma_2014} was used as optimizer. The batch size for all systems was 32.

The performance of the two baseline systems and the proposed systems are listed in Table \ref{tab:result}. Comparing A-LSTM and conventional LSTM shows that the A-LSTM is able to outperform the conventional LSTM by 5.5\% in terms of MAF. Since the weighted pooling layer can see the hidden values from all time steps, this improvement is not from the benefit of seeing more time steps in higher layer. It leveraged the advantage of the flexible time dependency modeling capability of A-LSTM. This is especially useful in emotion recognition, because emotion is usually shown a state within a range of time steps rather than at a time step instantly. In this study, we have 256 neurons in the BLSTM (each direction has 128 neurons), so we only need add 256 parameters, which is the $W$ size, to achieve this improve. This cost can be ignored compared with about 600 k parameters of network.

\begin{table}
	\fontsize{6}{9}\selectfont
	\centering
	\newvspace
	\caption{The performance of baseline and proposed systems. MAF is macro average F-score. MAP is macro average precision.
	}
	\begin{tabular*}{0.6\columnwidth}{@{\extracolsep{\fill}}c||c|c|c}
		
		\hline
		\hline
		Approach & MAF & MAP & Accuracy \\
		\hline
		conventional LSTM & 43.8 & 64.3 & 52.7 \\
		mean LSTM & 43.5 & 64.3 & 52.8 \\
		advanced LSTM & \textbf{46.2} & \textbf{65.8} & \textbf{55.3} \\
		\hline
		\hline
	\end{tabular*}
\newvspace
	\label{tab:result}
\end{table}

The results also show that there is no improvement when we fixed the weights. Comparing mean LSTM and A-LSTM implies that learnable weights are better. Learning weights as a framework of data-driven assignment allows the model to make the assignment according to different situations. It is better because time dependency may vary at different time steps.

\newvspace
\subsection{Comparison between RNN and DNN Frameworks}
\newvspace
We also built a DNN with multi-task learning for comparison. The network has two parts, shared part and separate part. The former part is shared by all the tasks, which has 2 fully connected layers with 4096 RELU neurons per layer. The later part has 3 separate sub-networks respectively for 3 tasks. Each sub-network has 1 fully connected layers with 2048 RELU neurons. On top of that, there is a softmax layer for classification. The batch size was 32 and dropout rate was 0.5. The optimizer was \emph{stochastic gradients descending} (SGD). We used IS10 feature set extracted with openSMILE \cite{Eyben_2010_2} as input because it was suitable for the three tasks. IS10 was z-normalized based on the mean and variance from training part. We also used the tool of Focal \cite{nikobrummer_2017} to fuse the results from these two frameworks.

The results of the experiment are shown in Table \ref{tab:comparison}. It is shown that the RNN framework is about 23.2 \% worse than DNN framework. There are two reasons here. First, we have very limited data, which is only about 3200 training utterances. This amount may not train RNN framework sufficiently, especially training RNN is more difficult than DNN. Second, all the utterances were well segmented in IEMOCAP. It may not have long silence and pause as the situation in real world. The fusion result shows combining the two frameworks is better than either single one. It indicates that RNN framework can complement the DNN even with few training data. Besides, there are about 58 M parameters in DNN which is about 100 times as the one in RNN which means that RNN will have low hardware requirement when it is employed.

\begin{table}
	\fontsize{6}{9}\selectfont
	\centering
	\newvspace
	\caption{The comparison between DNN and RNN frameworks. ``IS10" is Interspeech 2010 feature set. ``Seq" is the sequential acoustic feature. ``RNN+DNN" is the fusion result.
	}
	\begin{tabular*}{0.75\columnwidth}{@{\extracolsep{\fill}}c|c||c|c|c}
		
		\hline
		\hline
		Approach & feature & MAF & MAP & Accuracy \\
		\hline
		DNN & IS10 & 56.9 & 66.8 & 58.2 \\
		RNN & Seq & 46.2 & 65.8 & 55.3 \\
		RNN+DNN & IS10+Seq & \textbf{58.2} & \textbf{69.6} & \textbf{58.7} \\
		\hline
		\hline
	\end{tabular*}
\newvspace
	\label{tab:comparison}
\end{table}

%
%

\imagevspace
\section{Conclusion and Future Work}
\label{sec:conclusion}
\newvspace
We proposed a new type of LSTM, A-LSTM, in this paper. This was a early study of A-LSTM. We applied it in the weighted pooling RNN for emotion recognition. It is shown that the A-LSTM can outperform the conventional LSTM under weighted pooling RNN framework with few extra parameters. The improvement leverages the advantage of flexible time dependency modeling capability in A-LSTM. Even though the weighted pooling RNN framework can not beat the state-of-the-art DNN framework on IEMOCAP, it can complement the DNN to achieve better performance. It also has the advantage in practical application in real world.


Future work is necessary to explore A-LSTM in other tasks. The idea of combining states at multiple time steps can also be extended to \emph{gated recurrent unit} (GRU) in the future. More data is also needed for training the RNN framework.

\balance
\bibliographystyle{IEEEbib}
\bibliography{reference_TAO}

\begin{thebibliography}{10}

\bibitem{hochreiter_1997}
S.~Hochreiter and J.~Schmidhuber,
\newblock ``Long short-term memory,''
\newblock {\em Neural computation}, vol. 9, no. 8, pp. 1735--1780, 1997.

\bibitem{Schuller_2010}
B.~Schuller, S.~Steidl, A.~Batliner, F.~Burkhardt, L.~Devillers, C.~Muller, and
  S.~Narayanan,
\newblock ``The {INTERSPEECH} 2010 paralinguistic challenge,''
\newblock in {\em Interspeech 2010}, Makuhari, Japan, September 2010, pp.
  2794--2797.

\bibitem{florian_2015}
F.~Eyben, K.R. Scherer, B.~Schuller, J.~Sundberg, E.~Andre, C.~Busso, L.Y.
  Devillers, J.~Epps, P.~Laukka, S.S. Narayanan, and K.P. Truong,
\newblock ``The geneva minimalistic acoustic parameter set (gemaps) for voice
  research and affective computing,''
\newblock {\em IEEE TRANSACTIONS ON AFFECTIVE COMPUTING}, vol. 7, no. 2, pp.
  190--202, 2015.

\bibitem{mirsamadi_2017}
S.~Mirsamadi, E.~Barsoum, and C.~Zhang,
\newblock ``Automatic speech emotion recognition using recurrent neural
  networks with local attention,''
\newblock in {\em 2017 IEEE International Conference on Acoustics, Speech and
  Signal Processing (ICASSP)}, New Orleans, U.S.A., Mar. 2017, IEEE, pp.
  2227--2231.

\bibitem{liu_2014}
G.~Liu and J.~H. Hansen,
\newblock ``Supra-segmental feature based speaker trait detection,''
\newblock in {\em Proc. Odyssey}, 2014.

\bibitem{snyder_2015}
D.~Snyder, D.~Garcia-Romero, and D.~Povey,
\newblock ``Time delay deep neural network-based universal background models
  for speaker recognition,''
\newblock in {\em 2015 IEEE Workshop on Automatic Speech Recognition and
  Understanding (ASRU)}, Arizona, USA, 2015, IEEE, pp. 92--97.

\bibitem{peddinti_2015}
V.~Peddinti, D.~Povey, and S.~Khudanpur,
\newblock ``A time delay neural network architecture for efficient modeling of
  long temporal contexts.,''
\newblock in {\em INTERSPEECH 2015}, Dresden, Germany, Sept. 2015, pp.
  3214--3218.

\bibitem{waibel_1989}
A.~Waibel, T.~Hanazawa, G.~Hinton, K.~Shikano, and K.~Lang,
\newblock ``Phoneme recognition using time-delay neural networks,''
\newblock {\em IEEE transactions on acoustics, speech, and signal processing},
  vol. 37, no. 3, pp. 328--339, 1989.

\bibitem{he_2016}
K.~He, X.~Zhang, S.~Ren, and J.~Sun,
\newblock ``Deep residual learning for image recognition,''
\newblock in {\em the IEEE conference on computer vision and pattern
  recognition}, Washington, USA, Jun. 2016, pp. 770--778.

\bibitem{srivastava_2015}
R.~Srivastava, K.~Greff, and J.~Schmidhuber,
\newblock ``Highway networks,''
\newblock {\em arXiv preprint arXiv:1505.00387}, 2015.

\bibitem{zhang_2016}
Y.~Zhang, G.~Chen, D.~Yu, K.~Yao, S.~Khudanpur, and J.~Glass,
\newblock ``Highway long short-term memory rnns for distant speech
  recognition,''
\newblock in {\em 2016 IEEE International Conference on Acoustics, Speech and
  Signal Processing (ICASSP)}, 2016, Mar. 2016, IEEE, pp. 5755--5759.

\bibitem{kim_2017}
J.~Kim, M.~El-Khamy, and J.~Lee,
\newblock ``Residual lstm: Design of a deep recurrent architecture for distant
  speech recognition,''
\newblock {\em arXiv preprint arXiv:1701.03360}, 2017.

\bibitem{zhang_2015}
S.~Zhang, C.~Liu, H.~Jiang, S.~Wei, L.~Dai, and Y.~Hu,
\newblock ``Feedforward sequential memory networks: A new structure to learn
  long-term dependency,''
\newblock {\em arXiv preprint arXiv:1512.08301}, 2015.

\bibitem{soltani_2016}
R.~Soltani and H.~Jiang,
\newblock ``Higher order recurrent neural networks,''
\newblock {\em arXiv preprint arXiv:1605.00064}, 2016.

\bibitem{wang_2016}
Y.~Wang and F.~Tian,
\newblock ``Recurrent residual learning for sequence classification.,''
\newblock in {\em EMNLP}, 2016, pp. 938--943.

\bibitem{xia_2017}
R.~Xia and Y.~Liu,
\newblock ``A multi-task learning framework for emotion recognition using 2d
  continuous space,''
\newblock {\em IEEE Transactions on Affective Computing}, vol. 8, no. 1, pp.
  3--14, 2017.

\bibitem{parthasarathy_2017_1}
S.~Parthasarathy and C.~Busso,
\newblock ``Jointly predicting arousal, valence and dominance with multi-task
  learning,''
\newblock in {\em INTERSPEECH}, Stockholm, Sweden, Aug. 2017.

\bibitem{Busso_2008_5}
C.~Busso, M.~Bulut, C.C. Lee, A.~Kazemzadeh, E.~Mower, S.~Kim, J.N. Chang,
  S.~Lee, and S.S. Narayanan,
\newblock ``{IEMOCAP}: Interactive emotional dyadic motion capture database,''
\newblock {\em Journal of Language Resources and Evaluation}, vol. 42, no. 4,
  pp. 335--359, December 2008.

\bibitem{bahdanau_2016}
Dzmitry D.~Bahdanau, J.~Chorowski, D.~Serdyuk, and Yoshua Y.~Bengio,
\newblock ``End-to-end attention-based large vocabulary speech recognition,''
\newblock in {\em 2016 IEEE International Conference on Acoustics, Speech and
  Signal Processing (ICASSP)}, Shanghai, China, Apr. 2016, IEEE, pp.
  4945--4949.

\bibitem{scikit-learn}
F.~Pedregosa, G.~Varoquaux, A.~Gramfort, V.~Michel, B.~Thirion, O.~Grisel,
  M.~Blondel, P.~Prettenhofer, R.~Weiss, V.~Dubourg, J.~Vanderplas, A.~Passos,
  D.~Cournapeau, M.~Brucher, M.~Perrot, and E.~Duchesnay,
\newblock ``Scikit-learn: Machine learning in {P}ython,''
\newblock {\em Journal of Machine Learning Research}, vol. 12, pp. 2825--2830,
  2011.

\bibitem{kingma_2014}
D.~Kingma and J.~Ba,
\newblock ``Adam: A method for stochastic optimization,''
\newblock {\em arXiv preprint arXiv:1412.6980}, 2014.

\bibitem{Eyben_2010_2}
F.~Eyben, M.~W\"{o}llmer, and B.~Schuller,
\newblock ``{OpenSMILE}: the {Munich} versatile and fast open-source audio
  feature extractor,''
\newblock in {\em ACM International conference on Multimedia (MM 2010)},
  Florence, Italy, October 2010, pp. 1459--1462.

\bibitem{nikobrummer_2017}
{nikobrummer},
\newblock ``Focal,'' https://sites.google.com/site/nikobrummer/focal, 2017,
\newblock Retrieved Aug 1st, 2017.

\end{thebibliography}

\end{document}